\documentclass[10pt,twocolumn]{IEEEtran}
\usepackage{amsmath, amssymb, graphicx, mathtools, comment, xcolor, booktabs, footmisc, threeparttable,bbm, float}
\usepackage{cite}
\usepackage{geometry}
\usepackage[colorlinks=true, 
            linkcolor=blue, 
            citecolor=blue, 
            urlcolor=blue]{hyperref}

\newtheorem{lem}{Lemma}

\newtheorem{assumption}{Assumption}
\newtheorem{rem}{Remark}

\usepackage{tcolorbox}
\tcbuselibrary{breakable}
\newtheorem{thm}{Theorem}

\DeclareMathOperator*{\argmin}{\arg\!\min}

\DeclareMathOperator{\Var}{Var}
\DeclareMathOperator{\Ex}{\mathbb{E}}
\DeclareMathOperator{\Erfc}{Erfc}

\title{Energy-Aware Routing to\\Large Reasoning Models}
\author{
Austin R.\ Ellis-Mohr,
Max Hartman,
and Lav R.\ Varshney\thanks{The authors are with the Department of Electrical and Computer Engineering, University of Illinois Urbana-Champaign, Urbana, IL, USA (email: \{austine4, maxh3, varshney\}@illinois.edu).}\thanks{Varshney is also with the AI Innovation Institute, Stony Brook University, Stony Brook, NY, USA (email: lav.varshney@stonybrook.edu).}
}

\begin{document}
\maketitle

\begin{abstract}
Large reasoning models (LRMs) have heterogeneous inference energy costs based on which model is used and how much it reasons. To reduce energy, it is important to choose the right LRM and operate it in the right way. As a result, the performance of systems that dispatch tasks to different individual LRMs depend on the balance between mean energy provisioning and stochastic fluctuations. The critical regime is the unique operating point at which neither auxiliary energy nor baseline energy is systematically wasted. Increasing baseline supply shifts the system toward persistent over-supply and baseline-energy waste, while reducing supply induces persistent reliance on auxiliary energy. Yet in this regime, performance remains volatility-limited and so a second-order characterization provides further insights that we develop. Here, performance is governed by how variability is absorbed across time, models, and execution choices. This perspective highlights variance-aware routing and dispatch as a principled design axis, and provides a theoretical basis for developing energy-aware model routing policies. We characterize routing behavior under a simple dispatch policy based on training- and inference-compute scaling laws for LRMs.
\end{abstract}

\section{Introduction}
Large artificial intelligence (AI) models have become widely used in the past few years, and recently large reasoning models (LRMs) have become prominent \cite{PlaatWVBVB2025, EllisMohr2026TheoryInferenceCompute}. Such models are often infeasible for organizations to host locally due to hardware constraints, leading to the growth of large data centers---so-called \emph{AI factories}---with massively parallel hardware to host an ever-growing number of models of varying complexity. These data centers expend a significant amount of energy to process model requests.

There has been recent interest in powering data centers with renewable energy \cite{AgarwalSNIBCSK2021}. However, sources such as solar and wind introduce significant variability in available energy, which rarely aligns with usage. To mitigate variability, Varaiya et al.\ considered various ways to optimize risk-limited dispatch of energy for numerous physical workloads \cite{Varaiya2011RiskLimitedDispatchEnergy}, which can be extended to informational workloads \cite{Varshney2014RiskLimitedDispatchKnowledge}. Here we aim to draw on the specific properties of LRMs in AI factories, where there is a possibility of routing tasks to particular LRMs \cite{ShnitzerOSSSSTY2024}. Notably, the use and energy requirements of AI models also fluctuates with time \cite{Luccioni2023bloom}. Moreover, AI models with varying capabilities have different inference energy costs \cite{JEGHAMAKEH2025, TRIPPPGN2024, HUSOMGSS2024}. Larger AI models tend to yield better performance due to training-compute scaling \cite{NayakV2026}, but this typically comes with higher inference-time energy use. Further, due to inference-compute scaling of LRMs, more time/energy of computation may provide higher-quality responses \cite{NayakV2026, EllisMohr2026TheoryInferenceCompute}. As such, there are two main dimensions of energy optimization: which LRM to route to and how long to run it for a given workload.

Previous work has studied reducing energy requirements for large-scale AI systems in several ways. The Clover system experimentally showed that routing to a mixture of low- and high-quality models can improve energy efficiency, while maintaining performance \cite{LISGT2023}. EcoServe focused on GPU and CPU usage optimization, specifically by exploiting underutilized host CPUs and dynamically scaling GPUs and CPUs \cite{LIHCFSG2025}. Moreover, there are many works that focus on optimizing AI models directly. FrugalGPT proposed the prompt adaption, LLM approximation, and LLM cascade strategies \cite{ChenZZ2024}. Model pruning and knowledge distillation techniques have also been used for model efficiency \cite{SunLBK2024, MURALIDKARANSJC2024, HsiehLYNF2024}, which in turn decrease the energy consumed. Although these types of approaches improve different aspects of AI factory efficiency, they typically study model inference efficiency and power system considerations in isolation. Therefore, this does not explicitly address deployment settings in which renewable energy availability, inference cost heterogeneity, and deadline constraints must be jointly considered.

In this work, we introduce a mathematically principled formulation of the energy-aware model routing problem. This has formal similarity to information-theoretic investigations of optimal packet scheduling in energy harvesting systems \cite{YANGS2012,UlukusYESZGH2015}, in which packets and harvested energy arrive randomly, and the goal is to minimize the time to send data packets. There are several second-order characterizations of energy-harvesting channels that characterize channel dispersion using Berry-Esseen forms of the central limit theorem \cite{MolavianJaziY2015}, but we study first-order and second-order characterizations for our problem directly using properties of Brownian motion. Also, in our setup, renewable energy can be augmented with non-renewable energy to meet the task constraints (i.e., time and accuracy). The routing policy assigns each task to a hosted model with the objective of minimizing auxiliary energy consumption, subject to the request’s constraints.

In addition to our key results on the first- and second-order characterizations of energy-aware routing to LRMs, we also provide deep connections to training-compute and inference-compute scaling laws that are empirically well-established and for which there are nascent theoretical explanations~\cite{Bahri2024explainingscaling,NayakV2026,EllisMohr2026TheoryInferenceCompute,Wadell2025FMChem}. Basing task dispatch on these scaling laws not only reduces the need for an energy-heavy dispatcher running a large AI model itself, but also provides practical guidance for dispatch policies in deployed AI factories.

\section{System Model and Problem Formulation}
\begin{figure*}[t]
    \centering \includegraphics[width=.69\linewidth]{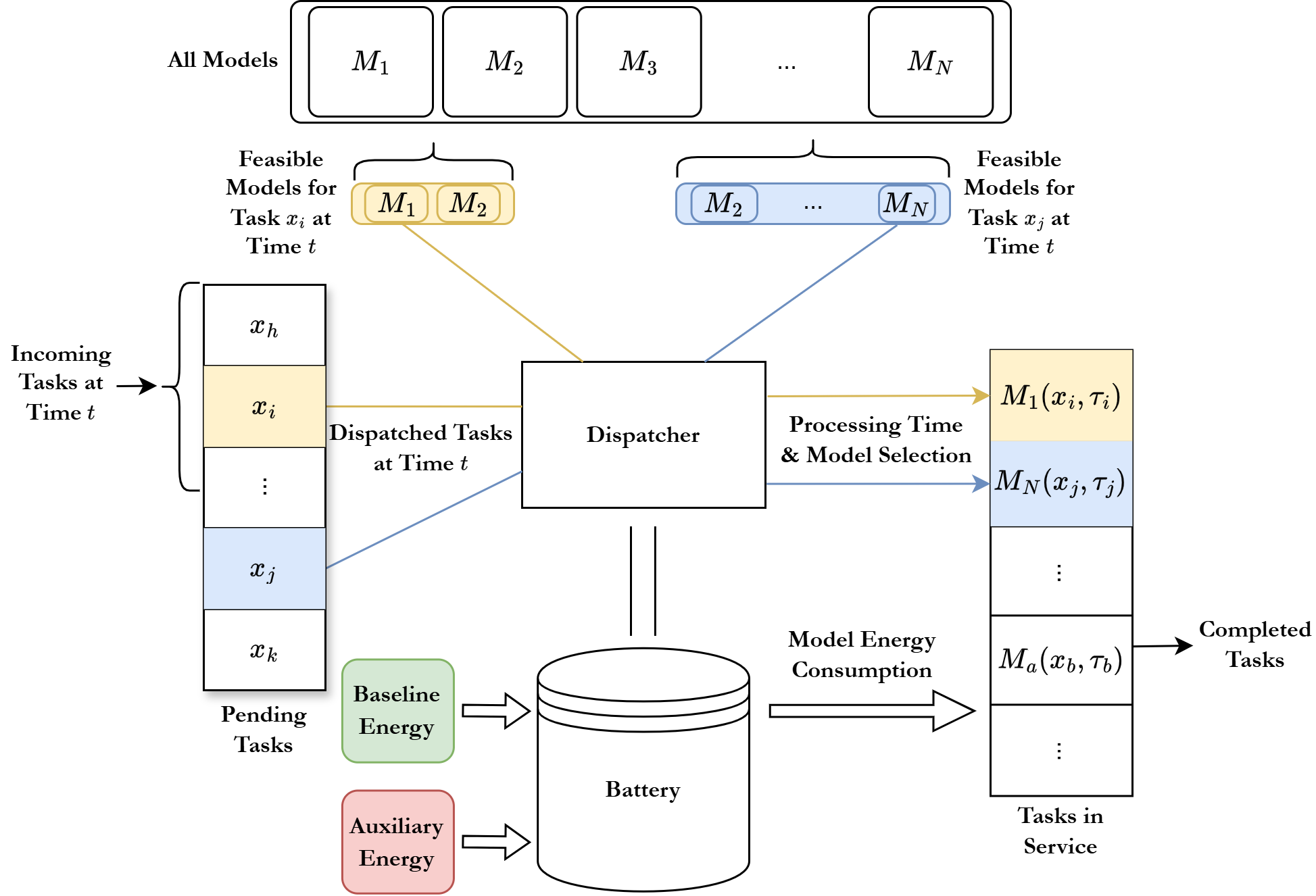}
    \caption{System diagram of the LRM routing system and energy flow: incoming tasks join the pending set, the dispatcher selects a feasible model and runtime for each dispatched task, and the battery supplies energy for execution, drawing auxiliary energy when needed.}
    \label{fig:system_diagram}
\end{figure*}

Consider an LRM routing system with unlimited parallel processing capacity: any number of tasks may be processed concurrently by any AI model, and the only binding resource is energy. The index set of available LRMs is $\mathcal{M}$. Tasks arrive stochastically and enter a router buffer (queue). At decision times, the router may dispatch any subset of queued tasks into processing (service). Any number of queued tasks may be dispatched to the same model $M_i\in\mathcal{M}$, even at the same time. Once dispatched, a task remains in processing for a policy-chosen thinking time and consumes energy throughout that interval; it is released (and stops using energy) at completion. Energy availability is also governed by stochastic processes $R$ for baseline energy and $G$ for auxiliary energy. See Fig.~\ref{fig:system_diagram}.

\subsection{Tasks, requirements, and arrivals}
A task is described by
\begin{equation}
    x \coloneqq (\theta, r)\in\mathcal{X},
\qquad
r \coloneqq (\lambda,\varepsilon),
\end{equation}
where $\theta\in\Theta$ is a (possibly vector-valued) task descriptor (e.g., difficulty and initial conditions such as context length), $\lambda>0$ is a latency deadline, and $\varepsilon\in(0,1)$ is an error tolerance. At time $t$ a random number $K_t$ of tasks arrive,
\[
\{x_{t,1},\dots,x_{t,K_t}\},
\]
drawn from an exogenous stochastic process. Each task $x$ has an arrival time $t_0(x)$ (the time it is generated and enters the queue).

\subsection{Router actions and task life-cycle}

A routing policy $\pi$ is online (nonanticipative): at each time $t$ it observes the current system state, including the stored energy and the set of tasks currently in the queue, and chooses: (i) which queued tasks to dispatch into processing at time $t$, and (ii) for each dispatched task, a model index and thinking time.

For any task $x$ with arrival time $t_0(x)$, let $s(x)\ge t_0(x)$ denote its dispatch time, i.e.\ the time at which the router launches the task into processing. At the dispatch time $s(x)$, the router selects a model index $i(x)\in\mathcal{M}$ and a thinking time $\tau(x)\ge 0$, yielding the per-task allocation
\begin{equation}
    (i(x),\tau(x)) = \pi(\text{state at time } s(x)).
\end{equation}
The task is in queue in interval $[t_0(x),\,s(x))$ and in service in interval $[s(x),\,s(x)+\tau(x))$. It completes at time $s(x)+\tau(x)$ and must satisfy the deadline constraint
\begin{equation}
s(x)+\tau(x) \le t_0(x)+\lambda(x).
\label{eq:deadline}
\end{equation}
Equivalently, a task cannot wait in the queue beyond its slack: at any dispatch time $s(x)$ the chosen $\tau(x)$ must fit in the remaining time-to-deadline $t_0(x)+\lambda(x)-s(x)$.

\subsection{Reliability model and oracle stopping}

When model $M_i$ processes task $x=(\theta,(\lambda,\varepsilon))$ for thinking time $\tau$, it succeeds, for a binary success variable, with probability
\begin{equation}
\mathbb{P}(\texttt{Success}=1 \mid x,i,\tau)=\psi_i(\theta;\tau),
\label{eq:psi}
\end{equation}
and the tolerance constraint is thus
\begin{equation}
\psi_i(\theta;\tau) \ge 1-\varepsilon.
\label{eq:phi_constraint}
\end{equation}
We adopt a best-case oracle stopping assumption: for any chosen $\tau$, the model runs until time $\tau$ and halts immediately when instructed. Thus, once the router dispatches a task and chooses $(i,\tau)$ satisfying~\eqref{eq:phi_constraint}, the task is guaranteed to stop (and therefore stop consuming energy) at completion time $s(x)+\tau$. This isolates routing/dispatch limits from early-stopping design. Termination can be implemented by enforcing a fixed compute budget while optionally triggering an earlier stop signal (for example by injecting a forced end-of-thinking token with reserved slack to finalize the response), with confidence- or verifier-based criteria as additional layers.

\subsection{System energetics}
If task $x=(\theta,(\lambda,\varepsilon))$ is dispatched to model $M_i$ for thinking time $\tau$, let $e_i(x,u)\ge 0$ denote its instantaneous energy-consumption rate at elapsed service time $u\in\{0,1,...,\tau-1\}$ (measured from the dispatch instant). The total energy required to run $x$ on $M_i$ for duration $\tau$
\begin{equation}
E_i(x,\tau) = \sum_{u=0}^{\tau-1} e_i(x,u).
\label{eq:Ei_def}
\end{equation}

Let $\mathcal{Q}(t)$ denote the set of tasks that have arrived by time $t$ and have not yet been dispatched. Under a policy $\pi$, each task $x$ has a dispatch time $s(x)\ge t_0(x)$ and (chosen at dispatch) an assigned model $i(x)\in\mathcal{M}$ and thinking time $\tau(x)\ge 0$, inducing a service interval $[s(x),\,s(x)+\tau(x))$. Define the set of tasks in service at time \(t\) as
\begin{equation}
    \mathcal{S}(t)\coloneqq \{x : t_0(x) \le t, s(x) \le t < s(x)+\tau(x) \}.
\end{equation}
The aggregate energy-consumption rate is then
\begin{equation}
C_t \coloneqq \sum_{x\in\mathcal{S}(t)} e_{i(x)} \bigl(x,\,t-s(x)\bigr),
\label{eq:c_total}
\end{equation}
and the stored energy evolves as
\begin{equation}
\widetilde B_{t+1}=\widetilde B_t+R_t+G_t-C_t,\qquad \widetilde B_t\ge 0\ \ \forall t,
\label{eq:Bnonnegdynamics}
\end{equation}
where $R_t\ge 0$ is the harvested baseline energy per step, $G_t\ge 0$ is the auxiliary energy rate, and $\widetilde{B}_0\ge0$ is the initial battery energy.

Our goal here is to characterize system-level feasibility and cost structures under baseline variability, while making the dependence on task requirements and routing decisions explicit. We adopt abstractions that isolate the dispatcher layer from other mechanisms. While modern inference engines support token- and iteration-level scheduling and sophisticated memory management \cite{Kwon2023PagedAttention}, we model each dispatched task as running continuously once started in order to focus on dispatcher-layer routing decisions without introducing additional engine-level coordination state. Accordingly, we abstract away mechanisms such as batching coordination across concurrent requests, queue reordering within the engine, and KV-cache management. Preemption may be incorporated by refining the time discretization and introducing micro-step scheduling (with cache and overhead modeling), which changes the set of feasible induced trajectories $\{C_t\}$ but not the form of the $\{R_t\}$--$\{C_t\}$ analysis provided.

\subsection{Objective and simplified dynamics}
A policy $\pi$ observes the arriving tasks and system state (including $\widetilde B_t$ and optionally harvest side-information) and makes online dispatch decisions, chooses allocations $(i(x),\tau(x))$ for dispatched tasks, and selects auxiliary energy usage $G_t$ when needed. In addition to observing realized harvests $\{R_t\}$, the dispatcher may observe exogenous covariates that capture predictable spatio-temporal structure in renewable availability. The goal is to satisfy, for every task, the deadline constraint~\eqref{eq:deadline} and the tolerance constraint~\eqref{eq:phi_constraint}, while minimizing reliance on auxiliary energy. Thus, considering the deficit at time $T$:
\begin{equation}
    D_T \coloneqq \sum_{t=0}^{T-1} G_t
\end{equation}
the objective is to find the policy that minimizes this deficit
\begin{equation}
    \min_{\pi}\ \Ex_{\pi} \left[D_T\right],
\end{equation}
subject to the corresponding energy dynamics~\eqref{eq:Bnonnegdynamics} and nonnegativity $\widetilde B_t\ge 0$ at all times.

The system dynamics may be cast without the nonnegativity constraint and auxiliary energy source as follows:
\begin{equation}
B_{t+1}=B_t+R_t-C_t.
\end{equation}
Furthermore, by Thm.~\ref{thm:objective}, proved in the Appendix, we can rewrite the objective in terms of these simplified dynamics since the cumulative injections of $G$ equal the maximal deficit of the unconstrained path on $B$.
\begin{thm}[Cumulative injections equal the maximal deficit of the unconstrained path]\label{thm:objective}\leavevmode

Fix a horizon $T\in\mathbb{N}$, and exogenous sequences $\{R_t\}_{t=0}^{T-1}$ and $\{C_t\}_{t=0}^{T-1}$ in $\mathbb{R}$. Let $t=0,1,\dots,T-1$, and let the \emph{uncontrolled} (possibly negative) battery trajectory $\{B_t\}_{t=0}^T$ be defined by
\begin{equation}
    B_{t+1}\coloneqq B_t + R_t - C_t.
\end{equation}
Let the \emph{controlled} battery trajectory $\{\widetilde B_t\}_{t=0}^T$ with nonnegative injections $\{G_t\}_{t=0}^{T-1}$ be defined by
\begin{align}
\widetilde B_{t+1} \coloneqq \widetilde B_t + R_t - C_t + G_t,\quad B_0 \ge 0.
\end{align}
where $G_t\ge 0$ for all $t$ with $\widetilde B_0 = B_0$. Consider the \emph{greedy} (minimal) choice
\begin{align}
G_t &\;\coloneqq \bigl(-(\widetilde B_t + R_t - C_t)\bigr)^+
\end{align}
where $(\cdot)^+=\max\{0,\cdot\}$, which is the smallest $G_t\ge 0$ that guarantees $\widetilde B_{t+1}\ge 0$ given $(\widetilde B_t,R_t,C_t)$.

Then the total injected energy satisfies
\begin{equation}
    D_T=\sum_{t=0}^{T-1} G_t=\Bigl(-\min_{0\le t\le T} B_t\Bigr)^+.
\end{equation}
\end{thm}
Thm.~\ref{thm:objective} is a pathwise identity for the harvested-energy sequence $\{R_t\}$ and the consumption sequence $\{C_t\}$; all routing-policy dependence enters through the induced consumption $C_t$ as defined in Eq.~\eqref{eq:c_total}. Then, the asymptotic objective with constraints is:
\begin{align} \label{eq:objective}
    \overline J^\star &\coloneqq \min_{\pi}\ \limsup_{T\to\infty}\ \mathbb{E}_{\pi}\!\left[\frac{1}{T}\Bigl(-\min_{0\le t\le T} B_t\Bigr)^+\right] \\ &\text{s.t. all task requirements are met.}\nonumber
\end{align}

\subsection{Feasible set}
For task $x = (\theta, (\lambda, \varepsilon))$ arriving at $t_0(x)$, define the minimum service time for model $M_n\in \mathcal{M}$:
\begin{equation}
    \tau_n^*(x) := \min\{\tau \geq 0 : \psi_n(\theta; \tau) \geq 1 - \varepsilon\}.\label{eq:toleranceconstraint}
\end{equation}
At time $t$, the remaining slack is $\sigma(x,t) \coloneqq t_0(x) + \lambda(x) - t$. Thus, the feasible model set at time $t$ is:
\begin{equation}
    \mathcal{M_F}(x, t) := \{n \in \mathcal{M} : \tau_n^*(x) \leq \sigma(x,t)\}.\label{eq:feasibleset}
\end{equation}
As $t$ increases, the remaining slack decreases, and $\mathcal{M_F}(x, t)$ shrinks as models drop out when their minimum service time exceeds remaining slack.

\section{Performance Analysis}
\label{sec:asymptotic}
We now characterize the fundamental limits of the routing system under ergodicity assumptions to demonstrate cost factors under an explicit myopic policy realization.

\subsection{Ergodic arrivals and harvesting}

\begin{assumption}\label{ass:stationary} We assume an empty initial battery, i.e., $B_0 = 0$. Task arrivals $\{(K_t,\{X_{t,k}\}_{k=1}^{K_t})\}_{t \ge 1}$ follow a Poisson process, where $K_t \sim \mathrm{Poisson}(\overline{K})$ and each task is represented by the random variable $X_{t,k}$. The task attributes $\{X_{t,k}\}$ are i.i.d.\ with distribution $f_X$. Energy harvests $\{R_t\}_{t \ge 1}$ form a mutually independent i.i.d.\ sequence with mean $\mathbb{E}[R_t] = \overline{R}$ and variance $\mathrm{Var}(R_t) = \sigma_R^2$. Moreover, the sequences $\{K_t\}$, $\{X_{t,k}\}$, and $\{R_t\}$ are mutually independent.

\end{assumption}

\begin{rem}[Interpretation of Assumption~\ref{ass:stationary}]
In practice, $R_t$ may be nonstationary or spatiotemporally correlated through exogenous covariates. Assumption~\ref{ass:stationary} is adopted to enable a tractable second-order (drift--fluctuation) analysis. Accordingly, our i.i.d.\ model can be viewed as a baseline for the short-timescale supply variability that remains after accounting for predictable structure (e.g., diurnal/seasonal components or location-dependent means via forecasting, planning, or workload shifting). We leave explicit correlated and location-aware energy models to future work.
\end{rem}

While finite-time scheduling and dispatch decisions may force the router to use a higher-energy allocation than the minimum available at a task's arrival, the arrival-feasible minimum energy provides a computable, policy-independent benchmark. The minimum energy to satisfy task $X$ on model $M_i$ under tolerance constraint~\eqref{eq:toleranceconstraint} is
\begin{equation}
    E_i^*(X)  \coloneqq  \sum_{u=0}^{\tau^*_i(X)-1} e_i(X,u).
\end{equation}
Let
\begin{equation}
    i_{\mathrm{LB}}(X)\in \argmin_{i\in \mathcal{M}_F(X,t_0(X))} E_i^*(X)
\end{equation}
be a (measurable) minimizer, and define $\tau_{\mathrm{LB}}(X)\coloneqq \tau^*_{i_{\mathrm{LB}}(X)}(X)$ and $e_{\mathrm{LB}}(X,u)\coloneqq e_{i_{\mathrm{LB}}(X)}(X,u)$ for $u=\{0,\dots,\tau_{\mathrm{LB}}(X)-1\}$, so that the lower bound on processing task $X$ is
\begin{equation}
    E_\mathrm{LB}(X)=E_{i_{\mathrm{LB}}(X)}^*(X)=\sum_{u=0}^{\tau_{\mathrm{LB}}(X)-1} e_{\mathrm{LB}}(X,u).
\end{equation}

The expected arrival-feasible energy lower bound per time step is the expected sum of per-task lower bounds over the random batch of tasks arriving in a slot:
\begin{equation}
    \overline{C}_\mathrm{LB}
    \coloneqq
    \Ex \Bigl[\sum_{k=1}^{K_t} E_\mathrm{LB}\!\left(X_{t,k}\right)\Bigr]
    = \overline K\,\Ex_{X}\!\left[E_\mathrm{LB}(X)\right],
\end{equation}
where the latter equality comes under Assumption~\ref{ass:stationary} and independence of $K_t$ and $\{X_{t,k}\}_{k=1}^{K_t}$. This provides a policy-independent lower bound on the long-run average energy consumption rate.

\subsection{Myopic dispatcher scaling analysis}

We now analyze a myopic baseline policy, $\pi_{\mathrm{my}}$, that dispatches each task in its arrival slot to the paired respective energy lower bound model. This does not exploit the option to slide service within latency windows; it is therefore a tractable reference model for variability and reserve scaling. We consider two myopic-in-time baselines that both begin service immediately upon arrival, but differ in how energy is accounted:
(i) lumped-at-arrival consumption and (ii) distributed-in-service consumption. Define the lumped myopic per-slot consumption:
\begin{equation}
    C_{t}^{\mathrm{my,lump}}
     \coloneqq 
    \sum_{k=1}^{K_t} E_{\mathrm{LB}}(X_{t,k}).
\end{equation}
Define the distributed myopic per-slot consumption as the total energy burned at slot $t$ by all tasks currently in service:
\begin{equation}
    C_t^{\mathrm{my,dist}}
     \coloneqq 
    \sum_{x \in \mathcal{S}(t)} e_{\mathrm{LB}}\bigl(x,\, t - s(x)\bigr).
\end{equation}
Lem.~\ref{lem:lump_pessimistic} in the Appendix shows that the lumped-at-arrival myopic model yields a pathwise upper bound on deficit for the corresponding distributed-in-service myopic model. We therefore use $C_t^{\mathrm{my,lump}}$ as a conservative reference process in the following scaling analysis. For clarity of notation, we drop the extra descriptor `$\mathrm{lump}$'.

\subsubsection{Battery random walk}
Analyzing how the auxiliary energy cost scales with time for the myopic dynamics yields a reference baseline for any proposed dispatcher policy. Under Assumption~\ref{ass:stationary}, the increments $R_t - C_t^{\mathrm{my}}$ are i.i.d., so the unconstrained battery process
\begin{equation}
    B_t^{\mathrm{my}} \;=\; \sum_{u=0}^{t-1} \bigl(R_u - C_u^{\mathrm{my}}\bigr)
\end{equation}
is a random walk. Its drift and per-step variance are characterized by the following results.

Since harvests are independent of task arrivals,
\begin{align}
    \Ex[B_t^{\mathrm{my}}]
    &= \Ex\Bigl[\sum_{u=0}^{t-1} R_u \Bigr]-\Ex\Bigl[\sum_{u=0}^{t-1} C_u^{\mathrm{my}} \Bigr],\\
    \Var(B_t^{\mathrm{my}})
    &= \Var\Bigl(\sum_{u=0}^{t-1} R_u\Bigr) + \Var\Bigl(\sum_{u=0}^{t-1} C_u^{\mathrm{my}}\Bigr).
\end{align}
The harvest contribution is $\Ex[\sum_{u=0}^{t-1} R_u] = t\overline{R}$ and $\Var(\sum_{u=0}^{t-1} R_u) = t\sigma_R^2$. Then, by stationarity, the expected cumulative consumption is
\begin{equation}
    \Ex\Bigl[\sum_{u=0}^{t-1} C_u^{\mathrm{my}} \Bigr]
    = t\,\Ex[C_u^{\mathrm{my}}]
    = t\,\overline{C}_\mathrm{LB},
\end{equation}
and by Lem.~\ref{lem:myopicexpenditurevariance} in the Appendix, the cumulative consumption variance is
\begin{align}
    \Var\Bigl(\sum_{u=0}^{t-1} C_u^{\mathrm{my}}\Bigr)
    &= t\,\Var(C_u^{\mathrm{my}})\\
    &= t\,\overline K\,\Ex_{X}\!\left[E_{\mathrm{LB}}(X)^2\right].
\end{align}
Thus, the lumped myopic battery process is described by the mean, $\mu_{B_{\mathrm{my}}}$, and variance, $\sigma_{B_{\mathrm{my}}}^2$:
\begin{equation}
    \mu_{B_{\mathrm{my}}}
    \coloneqq \tfrac{1}{t} \Ex[B_t^{\mathrm{my}}]
    = \overline{R}-\overline{C}_\mathrm{LB}
\end{equation}
\begin{equation}
    \sigma_{B_{\mathrm{my}}}^2
    \coloneqq \tfrac{1}{t}\Var(B_t^{\mathrm{my}}) = \sigma_R^2 + \overline{K}\Ex_{X}\!\left[E_{\mathrm{LB}}(X)^2\right].
\end{equation}

\subsubsection{Diffusion approximation}
Let $\{W_t\}_{t \ge 0}$ be standard Brownian motion. By Donsker's invariance principle~\cite{Donsker1951invariance}, the rescaled battery process converges in distribution as $T \to \infty$:
\begin{equation}
    \left\{\frac{B_{\lfloor uT \rfloor}^{\mathrm{my}} - \mu_{B_{\mathrm{my}}}\, uT}{\sigma_{B_{\mathrm{my}}}\sqrt{T}}\right\}_{u \in [0,1]}
    \;\Rightarrow\; \{W_u\}_{u \in [0,1]}.
    \label{eq:donsker}
\end{equation}
Here $u \in [0,1]$ is normalized time, with $u = t/T$ corresponding to real time $t \in [0, T]$. Rearranging~\eqref{eq:donsker} gives
\begin{equation}
    B_{\lfloor uT \rfloor}^{\mathrm{my}} \;\approx\; \mu_{B_{\mathrm{my}}}\, uT + \sigma_{B_{\mathrm{my}}}\sqrt{T}\, W_u.
    \label{eq:bm_approx_u}
\end{equation}
Substituting $t = uT$ and noting that $\sqrt{T}\, W_{t/T}$ is equal in distribution to $W_t$ by Brownian scaling yields
\begin{equation}
    B_t^{\mathrm{my}} \approx \mu_{B_{\mathrm{my}}}\, t + \sigma_{B_{\mathrm{my}}}\, W_t\sim \mathcal{N}\bigl(\mu_{B_{\mathrm{my}}}\, t,\; \sigma_{B_{\mathrm{my}}}^2\, t\bigr).
\end{equation}
Then, we show in Lem.~\ref{lem:reflectedrunningmin} and \eqref{eq:EDT_closed_form}, in the Appendix, the deficit $D_T^\mathrm{my}$ has a closed-form cumulative distribution function and expectation on its support $z \ge 0$.

In Thm.~\ref{thm:EDT_regimes}, proved in the Appendix, we analyze the expectation and prove that three simple regimes emerge under the large $T$ limit.
\begin{thm}[Expected deficit scaling across drift regimes]\label{thm:EDT_regimes}
Let $\mu\in\mathbb{R}$ and $\sigma>0$ be the drift and volatility parameters of a standard Brownian motion (see Lem.~\ref{lem:reflectedrunningmin}). Then the expected deficit satisfies, as $T\to\infty$,
\begin{equation}
\mathbb{E}[D_T]
=
\begin{cases}
|\mu|\,T + \dfrac{\sigma^2}{2|\mu|}, & \mu<0,\\
\dfrac{\sigma^2}{2\mu}, & \mu>0,\\
\sigma\sqrt{\dfrac{2T}{\pi}}, & \mu=0.
\end{cases}
\label{eq:EDT_regimes}
\end{equation}
In particular, the deficit grows linearly for $\mu<0$, remains bounded for $\mu>0$, and scales as $\sqrt{T}$ at $\mu=0$.
\end{thm}
For a persistent deficit $(\overline C_{\mathrm{LB}}>\overline R)$, the drift is negative and the running minimum is drift-dominated, so
$\mathbb{E}[D_T^\mathrm{my}]\approx|\mu_{B_{\mathrm{my}}}|T$. For a persistent surplus $(\overline C_{\mathrm{LB}}<\overline R)$, the drift is positive and the running minimum remains near its initial value, so
$\mathbb{E}[D_T^\mathrm{my}]=\mathcal{O}(1)$. But this corresponds to systematic over-generation, which is costly for real system design.

Thus, system designs may typically be tuned near balance where $\overline C_{\mathrm{LB}}\approx \overline R$, the running minimum is fluctuation-dominated. In particular, at zero drift,
\begin{equation}
    \mathbb{E}\!\left[D_T\right]
    \approx
    \sqrt{\frac{2}{\pi}}\;\sigma_{B_{\mathrm{my}}}\sqrt{T}.
    \label{eq:critical_myopic}
\end{equation}
Therefore, we turn our attention to this critical region of interest.

\subsection{Routing error as a function of fluctuations}

A routing policy $\pi$ selects, for each dispatched task $X$, a model–thinking-time pair $(i(X),\tau(X))$, possibly at random. Define the per-task excess energy when routing to model $i$ under policy $\pi$ as
\begin{equation}
    \Delta E^\pi_i(X) := E_i(X, \tau(X)) - E_{\mathrm{LB}}(X) \geq 0,
\end{equation}
with equality when $\pi$ achieves the arrival-feasible lower bound. Routing error induces per-task excess energy, which shifts the mean battery drift and accumulates linearly over the horizon. In contrast, stochastic supply–demand mismatch contributes a fluctuation-driven reserve cost that scales diffusively.

\begin{figure}[t]
    \centering
    \includegraphics[width=0.75\linewidth]{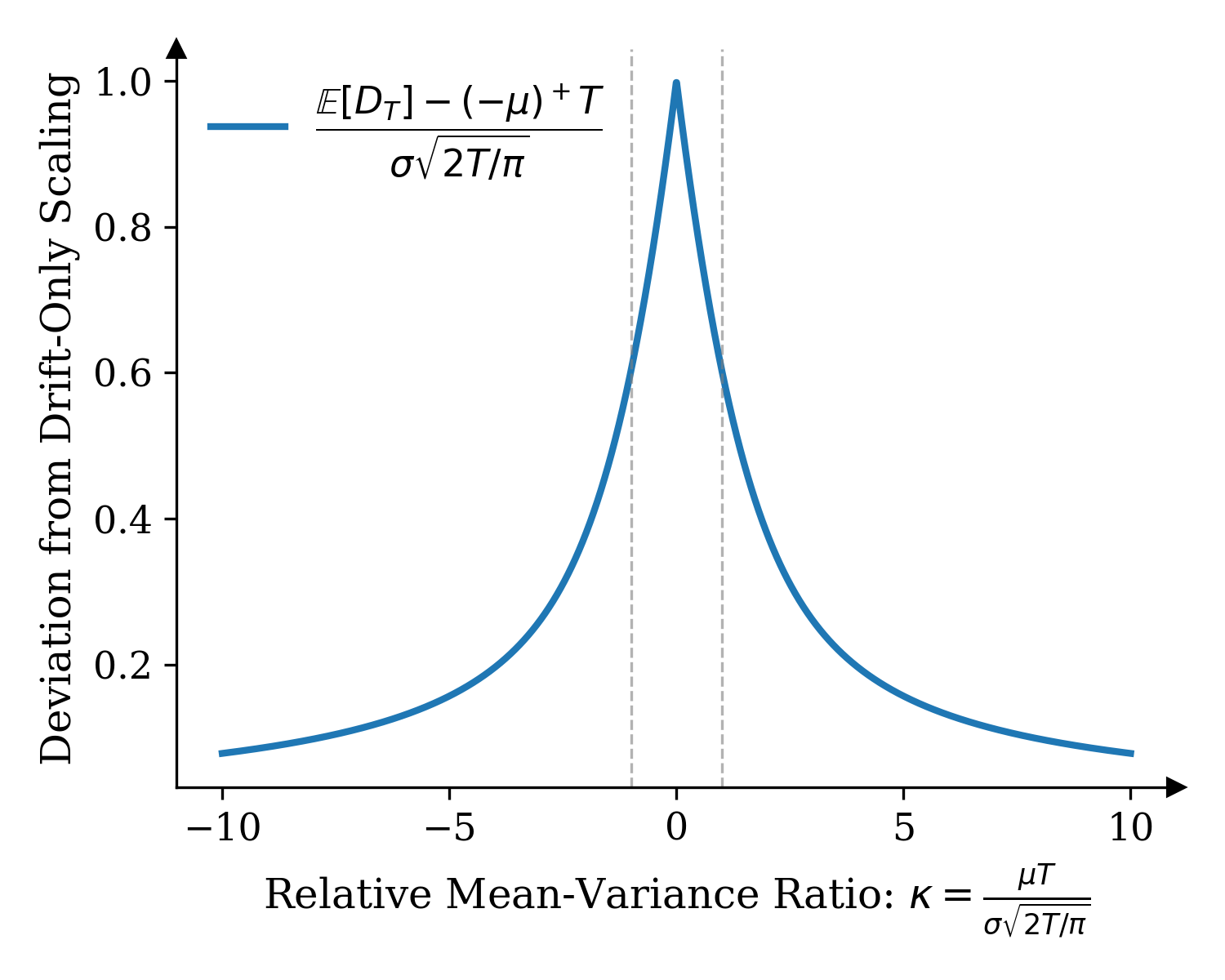}
    \caption{Deviation of the normalized expected reserve from drift-only scaling as a function of relative mean--variance ratio $\kappa=\mu T/(\sigma\sqrt{2T/\pi})$. The deviation is largest when drift and fluctuation contributions are of comparable scale and diminishes as drift dominance increases. Note that nonnegative drift denotes the surplus region.}
    \label{fig:fluctuation_deviation}
\end{figure}

To compare these effects on a common scale, we subtract the deterministic drift contribution $(-\mu)^+$ from the expected deficit $\Ex[D_T]$ and normalize by its zero-drift baseline $\sigma_{B_{\mathrm{my}}}\sqrt{2T/\pi}$. The resulting quantity $(\Ex[D_T]-(-\mu)^+ \ T)/(\sigma_{B_{\mathrm{my}}}\sqrt{2T/\pi})$ measures the finite-horizon deviation from drift-only scaling. Fig.~\ref{fig:fluctuation_deviation} plots this deviation as a function of the relative mean-variance ratio $\kappa=\mu T/(\sigma_{B_{\mathrm{my}}}\sqrt{2T/\pi})$. The deviation is largest when the linear drift and diffusive fluctuation terms are of comparable magnitude, and it decreases as the system moves into regimes dominated by either positive or negative drift. Thus, the figure highlights the parameter range in which stochastic fluctuations materially affect the reserve beyond what is predicted by mean drift alone.

This comparison clarifies the relative importance of routing accuracy and robustness. When the cumulative routing error
$\overline{K}\,\mathbb{E}_{\pi,X}[\Delta E_i^\pi(X)]\,T$
is significant relative to the fluctuation scale $\sigma_{B_{\mathrm{my}}}\sqrt{T}$, first-order drift dominates and improving routing accuracy yields the largest gains. When the two terms are comparable, fluctuation effects contribute non-negligibly to the reserve and policies may focus on accounting for variance in addition to mean optimality.

Significant differences in relative LRM energy efficiency on tasks motivate highly accurate and efficient task dispatching. However, the dispatcher itself consumes energy and time. Let $E_\pi^{\mathrm{self}}(X)$ and $\xi_\pi^{\mathrm{self}}(X)$ denote the energy and latency to process task $X$ and select a model, under policy $\pi$. The dispatcher latency reduces available slack, $\sigma(X, t) \to \sigma(X, t) - \xi_\pi^{\mathrm{self}}(X)$, potentially shrinking the feasible set $\mathcal{M}_F(X, t)$ and precluding lower-energy allocations. The total energy overhead per task is then $E_\pi^{\mathrm{self}}(X) + \Delta E_i^\pi(X)$: the cost of routing plus the excess from any suboptimal selection. A more capable router may reduce $\Ex[\Delta E_i^\pi(X)]$ but at the expense of increased $\Ex[E_\pi^{\mathrm{self}}(X)]$ and $\Ex[\xi_\pi^{\mathrm{self}}(X)]$. While this overhead is likely negligible relative to the difference between executing large models, it may become non-negligible if improved routing accuracy requires a higher-capacity learned router. This tradeoff motivates the use of explicit and accurate scaling laws to simplify the dispatch, reducing the dispatch-time burden while preserving accuracy.

\section{Compute Scaling}
\label{sec:scaling}
In the preceding analysis, we treated the per-task success behavior, completion time, and energy expenditure as known to the dispatcher. In practice, these quantities can be estimated using predictors that encode task difficulty and model response. To ground this assumption, we adopt empirical scaling laws for large transformer architectures \cite{Kaplan2020,Hoffmann2022} together with a recent theoretical extension for reasoning-style inference \cite{EllisMohr2026TheoryInferenceCompute}. In turn, the dispatch variables $(s(x), i(x), \tau(x))$ map these task- and model-level predictors into system-level completion and consumption dynamics (e.g., $C_t$ and the induced battery drift/variability) discussed under the preceding regime analysis. This makes the dependence of feasibility and energy on task requirements explicit.

\subsection{Token-based energy and latency}
When a task $X$ is dispatched to model $M_i$, the router allocates either a thinking time $\tau$ or, equivalently, an output-token budget $\Omega\in\mathbb{N}$. For analysis, we treat $\Omega$ as a continuous proxy and use fraktor symbols to denote the resulting continuous-time (or continuous-token) quantities; the discrete-time quantities used elsewhere in the paper are recovered by sampling at a chosen resolution.

We then take $E_{\mathrm{mem}}$ to be the energy cost per parameter memory access, $E_{\mathrm{comp}}$  the energy cost per floating-point operation, $n_{\mathrm{layers}}$  the number of transformer layers, and $d_{\mathrm{attn}}$  the attention hidden dimension. The model-dependent coefficients are then defined as $\alpha_i \coloneqq (E_{\mathrm{mem}}+2E_{\mathrm{comp}})\,N_i,$ $\beta_i \coloneqq E_{\mathrm{comp}}\,n_{\mathrm{layers},i}\,d_{\mathrm{attn},i},$ $a_i \coloneqq \frac{N_i}{\texttt{BW}}+\frac{2N_i}{\texttt{TP}},$ and $b_i \coloneqq \frac{n_{\mathrm{layers},i}\,d_{\mathrm{attn},i}}{\texttt{TP}}.$

Under the standard approximation that attention cost scales linearly with the current context length $L_{\mathrm{ctx}}$, the energy and time per generated token at context length $L_{\mathrm{ctx}}$ are modeled as:
\begin{equation}
\mathfrak{e}_{i,L_{\mathrm{ctx}}} = \alpha_i + 2\beta_i\,L_{\mathrm{ctx}}, \quad
    \xi_{i,L_{\mathrm{ctx}}} = a_i + 2b_i\,L_{\mathrm{ctx}},
\end{equation}
which correspond to parameter loading and feedforward/projection work (the constant terms) and attention over cached key-value pairs (the terms proportional to $L_{\mathrm{ctx}}$). While inference includes a prefill cost determined by the initial prompt, we omit this term for clarity and because our focus is on reasoning workloads where decode-time compute dominates and scales with the number of generated tokens. Even when decode is termwise linear and often memory-bandwidth bound due to streaming model weights and KV-cache reads, the cached context grows over long reasoning traces, so summing these per-token memory and compute costs over generated tokens produces an effective quadratic dependence on the total generation length. Then, we take $L_{\mathrm{ctx}}$ to grow proportionally with the number of generated tokens. Summing over $\Omega$ tokens yields the total energy and time:
\begin{equation}
\begin{aligned}
\mathfrak{E}_i(\Omega) &\coloneqq \sum_{v=0}^{\Omega-1} \mathfrak{e}_{i,\,v} \approx \alpha_i\,\Omega + \beta_i\,\Omega^2,\\
\mathfrak{T}_i(\Omega) &\coloneqq \sum_{v=0}^{\Omega-1} \xi_{i,\,v} \approx a_i\,\Omega + b_i\,\Omega^2.
\end{aligned}
\end{equation}
For a task $X$ with tolerance requirement $\varepsilon$, assume that accuracy is monotonically increasing in $\Omega$ and define the minimum token budget, $\Omega_i^*(X)$, for model $M_i$ as the solution to the tolerance constraint at equality:
\begin{equation}
    \psi_i(X;\Omega_i^*(X)) = 1-\varepsilon.
\end{equation}

\subsection{Discretization}
This induces a continuous completion time $\mathfrak{T}_i^*(X) \coloneqq \mathfrak{T}_i(\Omega_i^*(X)).$ We interface this continuous description with our discrete-time routing model by fixing a sampling resolution $\Delta>0$ (real time per slot). The induced discrete service time (in slots) is then $\tau_i^*(X) \coloneqq \left\lceil \mathfrak{T}_i^*(X)/\Delta\right\rceil,$
and feasibility under the discrete deadline constraint~\eqref{eq:deadline} defines the feasible set, $\mathcal{M}_F(X,t)$. The arrival-feasible minimum total energy for task $X$ on model $M_i \in \mathcal{M_F}(X,t_0)$ is $\mathfrak{E}_i^*(X) \coloneqq \mathfrak{E}_i(\Omega_i^*(X)).$

To obtain the discrete per-step energy profile $u\mapsto e_i(X,u)$ for our battery dynamics, define an auxiliary continuous time variable $\mathfrak{u}\in[0,\mathfrak{T}_i^*(X)]$. Since $\mathfrak{T}_i(\Omega)$ is strictly increasing for $\Omega\ge 0$, it admits an inverse. Letting $\Omega_i(\mathfrak{u})\coloneqq \mathfrak{T}_i^{-1}(\mathfrak{u})$ and $\mathfrak{E}_i(\mathfrak{u})\coloneqq \mathfrak{E}_i(\Omega_i(\mathfrak{u}))$, we define for $u=\{0,\dots,\tau_i^*(X)-1\},$
\begin{equation}
    e_i(X,u) \coloneqq \mathfrak{E}_i((u+1)\Delta)-\mathfrak{E}_i(u\Delta).
\end{equation}
By construction, $E_i^*(X)=\sum_{u=0}^{\tau_i^*(X)-1} e_i(X,u)$, allowing us to analyze routing for differing model sizes and capabilities under our mathematical framework.

\subsection{Scaling laws}
To provide an explicit, task-dependent expenditure induced by model size and inference scaling, we use the Directed Stochastic Skill Search inference scaling framework from \cite{EllisMohr2026TheoryInferenceCompute}. Following that work, we may regard the task descriptor as $\theta=(l,m)$, where $m\in\mathbb{N}$ is the number of sequential skills and $l\in\mathbb{R}_+$ parameterizes their difficulty. Then for chain-of-thought reasoning using $\Omega_i^*(X)$ tokens,
\begin{equation}
    \psi_i(X;\Omega_i^*(X))=I_{\mathfrak{p}_i(l)}(m,\Omega_i^*/\omega-m+1),
\end{equation}
where $I_x(a,b)$ is the regularized incomplete beta function, $\mathfrak{p}_i(l)\in[0,1]$ parameterizes the capability of model $i$ to reason about a task of some difficulty (with one being the highest success rate), and $\omega$ is a scaling factor for the number of tokens per skill.

To realize $\mathfrak{p}_i$ as a function of model size, we adopt training-compute optimal scaling as discussed by Hoffmann et al.~\cite{Hoffmann2022}. Selecting dataset size along the training-compute-optimal frontier, the token-level pretraining loss may be written as $\mathcal{L}_i(N_i)=\mathcal{L}_{\mathrm{irr}}+\Gamma N_i^{-\gamma}$ where $\mathcal{L}_\mathrm{irr}$ is an irreducible loss and $\Gamma, \gamma$ are based on model family efficiency for the given dataset. Then one simple functional form to model capability for a model of size $N_i$ is a sigmoid such that more difficult tasks (lower $l$) have a lower probability of success and vice versa:
\begin{equation}
\mathfrak{p}_i(l)=\mathrm{Sigmoid}\bigl(\mathfrak{b}(l-\mathcal{L}_i(N_i))\bigr),
\end{equation}
with $\mathfrak{b}$ parameterizing the steepness.

\subsection{Numerical simulation}
\begin{figure}
    \centering
    \includegraphics[width=0.75\linewidth]{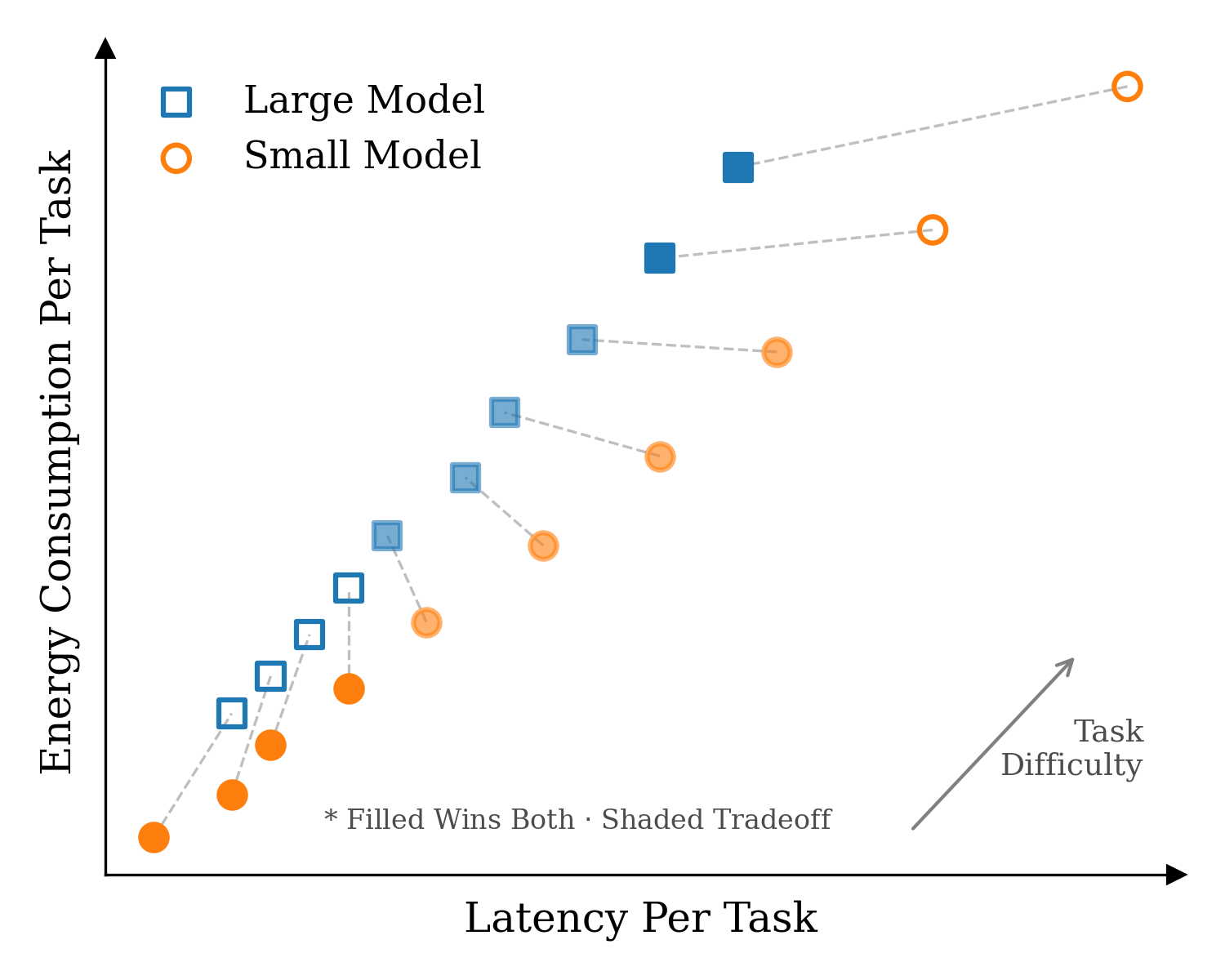}
    \caption{Energy consumption and latency per task to generate a response within an error tolerance for a large and small model. The small model is characterized by less-accurate, fast token generation, and the large model is characterized by more-accurate, slow token generation. For simpler tasks, the small model takes less time and energy, however as the task difficulty increases, the small model uses more energy and takes a longer amount of time before generating a correct response leading to a tradeoff before the large model becomes preferred.}
    \label{fig:consumption_vs_latency}
\end{figure}

\begin{figure}
    \centering
    \includegraphics[width=0.75\linewidth]{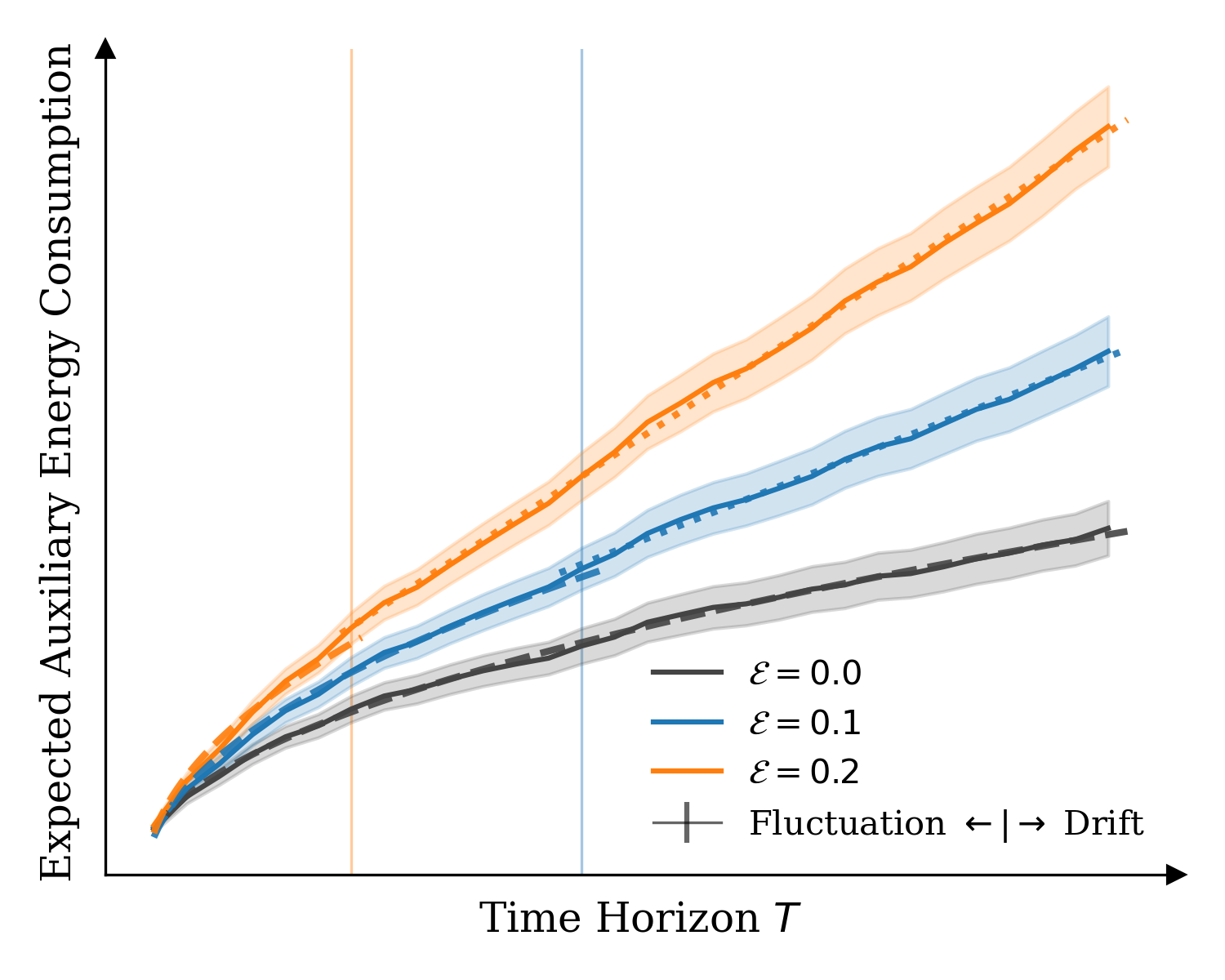}
    \caption{Expected auxiliary energy consumption $\Ex[D_T]$ versus time horizon $T$ under varying prediction errors $\mathcal{E}$ for the myopic policy. The zero error policy ($\mathcal{E} = 0$) exhibits square-root scaling throughout (dashed fit), while nonzero error policies transition from fluctuation-dominated (square-root scaling, dashed) to drift-dominated (linear scaling, dotted) regimes at the vertical markers. Shaded regions indicate standard error over 100 trials.}
    \label{fig:auxiliarycostvstime}
\end{figure}

Fig.~\ref{fig:consumption_vs_latency} illustrates the energy--latency tradeoff between two models of different sizes across varying task difficulties. For this realization, on easier tasks, the small model completes faster and consumes less energy. As difficulty increases, however, the small model requires substantially more tokens to meet the error tolerance, eventually crossing into a regime where the large model dominates on both axes. In between, there is a tradeoff as the smaller model takes longer but consumes less energy than the larger. This behavior is a core motivation for energy-aware routing: optimal dispatch requires matching task difficulty and requirements to model capability, and misrouting can incur significant excess energy $\Delta E_i^\pi(X)$ as illustrated in Fig.~\ref{fig:auxiliarycostvstime}.

To analyze how such routing errors propagate to system-level auxiliary costs, we introduce a prediction error $\mathcal{E} \in [0,1]$ controlling the probability of suboptimal model selection. When $\mathcal{E} > 0$, the dispatcher occasionally routes tasks to a higher-energy allocation than necessary, inducing nonzero expected excess $\mathbb{E}[\Delta E_i^\pi(X)] > 0$ and shifting the mean battery drift away from criticality.

Fig.~\ref{fig:auxiliarycostvstime} plots expected auxiliary energy $\mathbb{E}[D_T]$ against horizon $T$ for several values of $\mathcal{E}$. Simulation details are provided in Appendix~\ref{ap:simulationdetails}. The zero-error policy ($\mathcal{E} = 0$) remains critical and exhibits the $\sqrt{T}$ scaling of Thm.~\ref{thm:EDT_regimes}. Nonzero-error policies initially follow the same square-root trajectory while fluctuations dominate, but transition to linear scaling once cumulative drift $|\mu|T$ overtakes the diffusive term $\sigma\sqrt{T}$. The vertical markers indicate detected regime transitions, corresponding to the crossover region considered previously in Fig.~\ref{fig:fluctuation_deviation} where $|\kappa|$ grows large. This agrees with the theoretical prediction: routing errors are first-order effects that can eventually dominate the second-order fluctuation costs, underscoring the dual value of accurate, efficient dispatch combined with online policies to address fluctuations.

\section{Discussion}
Analyses in Sec.~\ref{sec:asymptotic}--\ref{sec:scaling} provide first- and second-order characterizations for the energy-aware LRM routing problem and further include LRM scaling laws (both training-compute and inference-compute). As shown, the specific selection of dispatcher policy is critical to minimizing auxiliary energy usage.

Future work may consider alternative dispatch policies. Note the formal similarities between the present problem and joint routing-scheduling in energy-harvesting communication networks \cite{Calvo-FullanaAMR2018}, for which backpressure-type algorithms \cite{TassiulasE1990} are developed. In our setting, a backpressure policy would route tasks by treating the request queue and the battery's energy deficit as competing pressures. This algorithm would aggressively use large, energy-intensive models to clear backlogs when renewable energy is abundant, while using efficient models during energy shortages.

The computation graphs of common inference-compute scaling techniques are tree-structured, but there may be settings with more general directed acyclic graphs that impose more complicated precedence structures, cf.~\cite{LechowiczSBHWD2025}. Our current work treats LRM tasks as independent from one another, but future work may consider dependencies that add more constraints while simultaneously enabling new opportunities for energy optimization. Renewable supply is often temporally structured and correlated (e.g., wind and solar), and prior work has emphasized the importance of accounting for such temporal correlation in sustainable system design~\cite{Gupta2022CarbonModeling}. Such correlation can shift reserve and storage requirements.

At the same time, many correlations are partly predictable (e.g., diurnal/seasonal structure and site-dependent means) and can be mitigated operationally by shifting flexible workloads across time and, when available, across geographically distributed sites~\cite{Colangelo2025AIGrid}. After accounting for this predictable structure, we conjecture the remaining short-timescale mismatch is closer to the i.i.d.\ baseline used here. We use the i.i.d.\ model primarily to obtain closed-form scaling and a clear baseline that clarifies the drift–fluctuation terms. Incorporating correlated and location-dependent energy models is an important direction for future work. Studying limited-capacity or leaky battery models also presents further avenues for exploration.

Furthermore, we note that currently we treat the  tolerance constraint $\varepsilon$ in \eqref{eq:phi_constraint} as deterministic. But, there may be settings where it may be stochastic as part of contracting or terms of use mechanisms for LRMs. Indeed, \cite{Varaiya2011RiskLimitedDispatchEnergy, Varshney2014RiskLimitedDispatchKnowledge} introduce a probability distribution on risk tolerance and construct contract mechanisms with tranches of different risks.

Overall, this work introduces the mathematical problem of energy-aware routing for large reasoning models. The general framework enables a variety of extensions.

\bibliographystyle{IEEEtran}
\bibliography{abrv,conf_abrv,arem_lib,project_lib}

\begin{appendices}
\section{Supporting Results and Proofs}
\begin{IEEEproof}[Proof of Theorem~\ref{thm:objective}]
Define the cumulative injected energy up to time $t$ by
\begin{equation}
    S_t \coloneqq \sum\nolimits_{s=0}^{t-1} G_s,
\qquad t=0,1,\dots,T,
\end{equation}
with the convention $S_0=0$. Summing the controlled recursion and comparing to the uncontrolled one yields the fundamental identity
\begin{equation}\label{eq:tildeB_equals_B_plus_S}
\widetilde B_t \;=\; B_t + S_t,
\qquad t=0,1,\dots,T.
\end{equation}
Indeed, both processes have the same initial condition $\widetilde B_0=B_0\ge 0$, and
\begin{equation}
    \widetilde B_{t+1}-B_{t+1} = (\widetilde B_t-B_t) + G_t,
\end{equation}
so by induction $\widetilde B_t-B_t=\sum_{s=0}^{t-1}G_s=S_t$.

Now define the running minimum of the uncontrolled trajectory and its associated deficit:
\begin{equation}
    \mu_t \coloneqq \min_{0\le u\le t} B_u,\quad D_t \coloneqq (-\mu_t)^+.
\end{equation}
We will show that under the greedy policy,
\begin{equation}\label{eq:claim_S_equals_D}
S_t = D_t \quad \text{for all } t=0,1,\dots,T,
\end{equation}
which immediately implies the theorem by taking $t=T$.

\medskip
\noindent\textbf{Step 1: Greedy update for $S_{t+1}$.}
Using $\widetilde B_t = B_t + S_t$ from \eqref{eq:tildeB_equals_B_plus_S} and the greedy definition,
\begin{align}
  G_t &= \bigl(-(\widetilde B_t + R_t - C_t)\bigr)^+\\
&= \bigl(-(B_t+S_t+R_t-C_t)\bigr)^+.  
\end{align}
But $B_{t+1}=B_t+R_t-C_t$, hence
\begin{equation}\label{eq:G_t_in_terms_of_B_and_S}
G_t = \bigl(-(B_{t+1}+S_t)\bigr)^+.
\end{equation}
Therefore,
\begin{align}\label{eq:S_update}
S_{t+1} &= S_t + G_t= S_t + \bigl(-(B_{t+1}+S_t)\bigr)^+\\
&= \max\{S_t,\,-B_{t+1}\}.
\end{align}
The last equality follows by a case split: if $B_{t+1}+S_t\ge 0$ then $G_t=0$ so $S_{t+1}=S_t$; otherwise $G_t=-(B_{t+1}+S_t)$ so $S_{t+1}=-B_{t+1}$.

\medskip
\noindent\textbf{Step 2: Induction that $S_t=D_t$.}
We prove \eqref{eq:claim_S_equals_D} by induction on $t$.

\emph{Base case ($t=0$):} $S_0=0$. Also $\mu_0=\min\{B_0\}=B_0\ge 0$, so $D_0=(-\mu_0)^+=0$. Hence $S_0=D_0$.

\emph{Inductive step:} Assume $S_t=D_t$ for some $t\in\{0,1,\dots,T-1\}$. Then by \eqref{eq:S_update},
\begin{equation}
   S_{t+1} = \max\{S_t,\,-B_{t+1}\} = \max\{D_t,\,-B_{t+1}\}. 
\end{equation}
On the other hand, since $\mu_{t+1}=\min\{\mu_t,B_{t+1}\}$,
\begin{align}
D_{t+1} &= \bigl(-\mu_{t+1}\bigr)^+= \Bigl(-\min\{\mu_t,B_{t+1}\}\Bigr)^+\\
&= \max\Bigl\{(-\mu_t)^+,\; (-B_{t+1})^+\Bigr\}\\
&= \max\{D_t,\,-B_{t+1}\},
\end{align}
where in the last equality we used $D_t\ge 0$, so $\max\{D_t,(-B_{t+1})^+\}=\max\{D_t,-B_{t+1}\}$.
Thus $S_{t+1}=D_{t+1}$, completing the induction.

Therefore \eqref{eq:claim_S_equals_D} holds for all $t$, and in particular
\begin{equation}
    \sum\nolimits_{t=0}^{T-1} G_t = S_T = D_T = \Bigl(-\min_{0\le t\le T} B_t\Bigr)^+.
\end{equation}
This is exactly the desired identity.
\end{IEEEproof}

\begin{lem}[Lumped myopic is pathwise pessimistic]\label{lem:lump_pessimistic}
For every sample path and all $t \ge 0$,
\begin{equation}
    B_t^{\mathrm{my,dist}} \;\ge\; B_t^{\mathrm{my,lump}}.
\end{equation}
Consequently, for every horizon $T$,
\begin{equation}
    \bigl(-\min_{0 \le t \le T} B_t^{\mathrm{my,dist}}\bigr)^+
    \;\le\;
    \bigl(-\min_{0 \le t \le T} B_t^{\mathrm{my,lump}}\bigr)^+.
\end{equation}
\end{lem}

\begin{IEEEproof}[Proof of Lemma~\ref{lem:lump_pessimistic}]
Fix a sample path. Under the distributed model, the cumulative energy consumed by task $x$ through the end of slot $t \ge s(x)$ is
\begin{align}
    \sum_{u=0}^{(t - s(x)) \wedge (\tau(x)-1)}\!\!\!\! e_{\mathrm{LB}}(x, u)
\;&\le\;
\sum_{u=0}^{\tau(x)-1} e_{\mathrm{LB}}(x, u)\\
\;&=\;
E_{\mathrm{LB}}(x),
\end{align}
with equality only once the task completes. Summing over all tasks that have arrived by time $t$, the cumulative distributed consumption is at most the cumulative lumped consumption. Since both battery processes share the same initial level $B_0$ and harvest sequence $\{R_t\}$, subtracting a smaller cumulative consumption yields $B_t^{\mathrm{my,dist}} \ge B_t^{\mathrm{my,lump}}$ for all $t$. Taking minima and applying $(\cdot)^+$ preserves the inequality.
\end{IEEEproof}

\begin{lem}[Variance of arrival-feasible consumption]\label{lem:myopicexpenditurevariance}
Under Assumption~\ref{ass:stationary} with Poisson arrivals,
\begin{equation}
    \Var\Bigl(\sum\nolimits_{t=0}^{T-1} C_t^\mathrm{MY}\Bigr) = \overline{K}T \cdot \Ex\left[E_{\mathrm{LB}}(X)^2\right].
\end{equation}
\end{lem}
\begin{IEEEproof}[Proof of Lemma~\ref{lem:myopicexpenditurevariance}]
Let $N_T = \sum_{t=0}^{T-1} K_t$ denote the total arrivals in $[0, T-1]$, and let $S = \sum_{t=0}^{T-1} C_t^\mathrm{my,lump} = \sum_{t=0}^{T-1} \sum_{k=1}^{K_t} E_{\mathrm{LB}}(X_{t,k})$, which is a sum of $N_T$ i.i.d.\ terms. By the law of total variance,
\begin{equation}
    \Var(S) = \Ex[\Var(S \mid N_T)] + \Var(\Ex[S \mid N_T]).
\end{equation}
Conditioning on $N_T = n$, the sum $S$ comprises $n$ i.i.d.\ copies of $E_{\mathrm{LB}}(X)$. For the conditional expectation,
\begin{equation}
    \Ex[S \mid N_T = n] = n \cdot \Ex[E_{\mathrm{LB}}(X)].
\end{equation}
For the conditional variance, independence of the $E_{\mathrm{LB}}(X_{t,k})$ given $N_T = n$ yields
\begin{align}
    \Var & (S \mid N_T = n) = \Var\Bigl(\sum_{i=0}^{n-1} E_{\mathrm{LB}}(X_i)\Bigr)\\ &= \sum_{i=0}^{n-1} \Var(E_{\mathrm{LB}}(X_i))= n \cdot \Var(E_{\mathrm{LB}}(X)).
\end{align}
Substituting,
\begin{align}
    \Var(S) &= \Ex[N_T] \cdot \Var(E_{\mathrm{LB}}(X))\nonumber\\ &+ \Var(N_T) \cdot \left(\Ex[E_{\mathrm{LB}}(X)]\right)^2.
\end{align}
Since sums of independent Poissons are Poisson, $\Ex[N_T] = \Var(N_T) = \overline{K}T$, yielding
\begin{align}
    \Var(S) &= \overline{K}T \bigl(\Var(E_{\mathrm{LB}}(X)) + \left(\Ex[E_{\mathrm{LB}}(X)]\right)^2\bigr) \\
    &= \overline{K}T \cdot \Ex\left[E_{\mathrm{LB}}(X)^2\right].
\end{align}
\end{IEEEproof}

\begin{lem}[Running minimum of Brownian motion with drift]\label{lem:reflectedrunningmin}
Let $\{B_t\}_{t\ge 0}$ be Brownian motion with drift $\mu$ and volatility $\sigma>0$, i.e.,
$B_t=\mu t+\sigma W_t$ where $\{W_t\}_{t\ge 0}$ is standard Brownian motion. Define the deficit
\begin{equation}
    D_T \coloneqq \Bigl(-\min_{0\le t\le T} B_t\Bigr)^+.
\end{equation}
Then for $z\ge 0$,
\begin{equation}
\mathbb{P}(D_T\le z)
=\Phi\!\left(\!\frac{z+\mu T}{\sigma\sqrt{T}}\right)
- e^{\frac{-2\mu z}{\sigma^2}}\!
  \Phi\!\left(\!\frac{\mu T-z}{\sigma\sqrt{T}}\right),
\label{eq:deficit_dist}
\end{equation}
where $\Phi(u)\coloneqq \frac{1}{\sqrt{2\pi}}\int_{-\infty}^{u}e^{-s^2/2}\,ds$ is the standard normal CDF.
\end{lem}

\begin{IEEEproof}[Proof of Lemma~\ref{lem:reflectedrunningmin}]
Fix $T>0$ and let $m_T\coloneqq \min_{0\le t\le T}B_t$, which exists almost surely since $B$ has continuous
sample paths on the compact interval $[0,T]$. For $z\ge 0$,
\begin{align}
\mathbb{P}(D_T\le z)
&=\mathbb{P}\!\left(\bigl(-m_T\bigr)^+\le z\right)\\
=\mathbb{P}&\left(-m_T\le z\right)=\mathbb{P}\!\left(m_T\ge -z\right).
\label{eq:DT_to_min}
\end{align}
The $(\cdot)^+$ is redundant for $z\ge 0$. Define the rescaled process $Y_t \coloneqq B_t/\sigma = W_t + \tilde{\mu} t$, where $\tilde{\mu} \coloneqq \mu/\sigma$.
Then $m_T = \sigma \min_{0 \le t \le T} Y_t$, and for $z \ge 0$,
\begin{align}
\mathbb{P}(m_T \ge -z)
&= \mathbb{P} \Bigl(\inf_{0 \le t \le T} Y_t \ge -\frac{z}{\sigma}\Bigr)\\
&= 1 - \mathbb{P} \Bigl(\inf_{0 \le t \le T} Y_t \le -\frac{z}{\sigma}\Bigr),
\label{eq:scaling_step}
\end{align}
where we used continuity of $Y$ to identify $\min=\inf$ and to note that strict versus non-strict inequalities at the boundary are immaterial.

The process $Y_t=W_t+\tilde\mu t$ is Brownian motion with drift $\tilde\mu$ and unit volatility, denoted
$W^{(\tilde\mu)}_t$ by Borodin and Salminen \cite{Borodin2002Handbook}. Indeed by \cite[formula 1.2.4 (p.~257)]{Borodin2002Handbook}, for drift parameter
$\alpha\in\mathbb{R}$ and level $y\le x$,
\begin{align}
\label{eq:bs_formula}
\mathbb{P}_x \Bigl(\inf_{0\le s\le t} W^{(\alpha)}_s \le y\Bigr)
&= \tfrac{1}{2}\Erfc\!\left(\tfrac{x-y+\alpha t}{\sqrt{2t}}\right) \nonumber\\
+& \tfrac{1}{2}e^{2\alpha(y-x)}\Erfc\!\left(\tfrac{x-y-\alpha t}{\sqrt{2t}}\right)
\end{align}
where $\Erfc(u)\coloneqq \frac{2}{\sqrt{\pi}}\int_{u}^{\infty}e^{-r^2}\,dr$ is the complementary error function.

Apply \eqref{eq:bs_formula} with $x=0$, $y=-z/\sigma\le 0$, $t=T$, and $\alpha=\tilde\mu=\mu/\sigma$:
\begin{align}
\mathbb{P} \Bigl(\inf_{0\le s\le T}Y_s \le -&\frac{z}{\sigma}\Bigr)
=\frac{1}{2} \Erfc\!\left(\frac{z+\mu T}{\sigma\sqrt{2T}}\right)\nonumber\\
&+\frac{1}{2} e^{-2\mu z/\sigma^2}\Erfc\!\left(\frac{z-\mu T}{\sigma\sqrt{2T}}\right).
\label{eq:infY_erfc}
\end{align}

Use the standard identity, valid for all $u\in\mathbb{R}$,
\begin{equation}
    \frac{1}{2}\Erfc\!\left(\frac{u}{\sqrt{2}}\right)=1-\Phi(u),
\end{equation}
together with $1-\Phi(a)=\Phi(-a)$. With $u_+\coloneqq \frac{z+\mu T}{\sigma\sqrt{T}}$ and
$u_-\coloneqq \frac{z-\mu T}{\sigma\sqrt{T}}$, \eqref{eq:infY_erfc} becomes
\begin{align}
\lefteqn{\mathbb{P} \Bigl(\inf_{0\le s\le T}Y_s \le -\frac{z}{\sigma}\Bigr)} \nonumber\\
&=\bigl(1-\Phi(u_+)\bigr)+e^{-2\mu z/\sigma^2}\bigl(1-\Phi(u_-)\bigr) \nonumber\\
&=\Phi(-u_+)+e^{-2\mu z/\sigma^2}\Phi(-u_-) \nonumber\\
\begin{split}
&=\Phi\!\left(\tfrac{-z-\mu T}{\sigma\sqrt{T}}\right) +e^{-2\mu z/\sigma^2}\Phi\!\left(\tfrac{\mu T-z}{\sigma\sqrt{T}}\right).
\end{split}
\label{eq:infY_phi}
\end{align}
Finally, combine \eqref{eq:DT_to_min}, \eqref{eq:scaling_step}, and \eqref{eq:infY_phi} to obtain
\begin{align}
    \mathbb{P}(D_T\le & z)
=1-\mathbb{P} \Bigl(\inf_{0\le s\le T}Y_s \le -\frac{z}{\sigma}\Bigr)\nonumber\\
&=\Phi\!\left(\frac{z\!+\!\mu T}{\sigma\sqrt{T}}\right) \!
- \!e^{-2\mu z/\sigma^2}\!\Phi\!\left(\frac{\mu T\!-\!z}{\sigma\sqrt{T}}\right),
\end{align}
which is \eqref{eq:deficit_dist}.
\end{IEEEproof}

\begin{IEEEproof}[Proof of Theorem~\ref{thm:EDT_regimes}]
Since $D_T\ge 0$ a.s., its CDF satisfies $\mathbb{P}(D_T\le z)=0$ for $z<0$, and
\begin{equation}
\mathbb{E}[D_T]=\int_{0}^{\infty}\mathbb{P}(D_T>z)\,dz.
\label{eq:tail_identity}
\end{equation}
From Lem.~\ref{lem:reflectedrunningmin}, for $z\ge 0$,
\begin{align}
    \mathbb{P}(D_T>z) &=
1-\Phi\!\left(\frac{z+\mu T}{\sigma\sqrt{T}}\right)\nonumber\\
&+e^{-2\mu z/\sigma^2}\,
\Phi\!\left(\frac{\mu T-z}{\sigma\sqrt{T}}\right).
\end{align}
Substitute into \eqref{eq:tail_identity} and change variables $z=\sigma\sqrt{T}\,b$, with
$a\coloneqq \mu\sqrt{T}/\sigma$, to obtain
\begin{align}
&\mathbb{E}[D_T]
\!=\!\sigma\sqrt{T}\int_{0}^{\infty}\!(Q(a\!+\!b)\!+\!e^{-2ab}\Phi(a\!-\!b))\,db,\\ 
&Q(u)\coloneqq 1-\Phi(u).
\label{eq:EDT_integral_ab}
\end{align}

The two integrals in \eqref{eq:EDT_integral_ab} admit standard closed forms:
\begin{equation}
\int_{0}^{\infty}Q(a+b)\,db =\phi(a)-aQ(a),\label{eq:integ1}
\end{equation}
\begin{equation}
    \int_{0}^{\infty}e^{-2ab}\Phi(a-b)\,db =\frac{2\Phi(a)-1}{2a},
\quad a\neq 0,
\label{eq:integ2}
\end{equation}
where $\phi(u)\coloneqq (2\pi)^{-1/2}e^{-u^2/2}$.
Substituting \eqref{eq:integ1} and  \eqref{eq:integ2} into \eqref{eq:EDT_integral_ab} yields, for $a=\frac{\mu\sqrt{T}}{\sigma}\neq 0$,
\begin{equation}
\mathbb{E}[D_T]
=\sigma\sqrt{T}\Bigl(\phi(a)-aQ(a)+\frac{2\Phi(a)-1}{2a}\Bigr).
\label{eq:EDT_closed_a}
\end{equation}
Equivalently, for $\mu\neq 0$,
\begin{equation}
\begin{aligned}
\mathbb{E}[D_T]
=
&\sigma\sqrt{T}\,\phi \bigl( \mu\sqrt{T}/\sigma \bigr)\\
-&\mu T \bigl(1-\Phi \bigl(\mu\sqrt{T}/\sigma \bigr)\bigr)\\
+&\frac{\sigma^2}{2\mu} \bigl(2\Phi \bigl(\mu\sqrt{T}/\sigma\bigr)-1\bigr).
\label{eq:EDT_closed_form}
\end{aligned}
\end{equation}

We now take limits.

\paragraph{Case $\mu>0$ ($a\to+\infty$)}
As $a\to+\infty$, $\phi(a)\to 0$, $Q(a)\to 0$, and $2\Phi(a)-1\to 1$, hence
\begin{equation}
    \mathbb{E}[D_T]=\sigma^2/(2\mu).
\end{equation}

\paragraph{Case $\mu<0$ ($a\to-\infty$)}
As $a\to-\infty$, $\phi(a)\to 0$, $Q(a)\to 1$, and $2\Phi(a)-1\to -1$, hence
\begin{equation}
    \mathbb{E}[D_T]
=-\mu T-\frac{\sigma^2}{2\mu}
=|\mu|T+\frac{\sigma^2}{2|\mu|}.
\end{equation}

\paragraph{Case $\mu=0$}
Take $\mu\to 0$ in \eqref{eq:EDT_closed_a}. Using $\phi(a)\to\phi(0)$,
$Q(a)\to\frac12$, and $\frac{2\Phi(a)-1}{2a}\to\phi(0)$, we obtain
\begin{equation}
    \mathbb{E}[D_T]\to \sigma\sqrt{T}\bigl(\phi(0)+\phi(0)\bigr)
=\sigma\sqrt{2T/\pi},
\end{equation}
which also holds at $\mu=0$ exactly.
This establishes \eqref{eq:EDT_regimes}.
\end{IEEEproof}

\section{Numerical Simulation Details}\label{ap:simulationdetails}
We evaluate a parametric inference--energy framework using abstract model, hardware, and task parameters (not deployed models), operating under Assumption~\ref{ass:stationary}. Two hosted model sizes are considered: $1$B and $10$B parameters, each with $n_{\mathrm{layers}}=48$ and attention dimension $d_{\mathrm{attn}}=2048$. The energy cost per parameter memory access is $E_{\mathrm{mem}}=10^{-11}$ J/parameter, and the energy cost per floating-point operation is $E_{\mathrm{comp}}=10^{-12}$ J/FLOP. Hardware parameters are fixed to memory bandwidth $BW=5\times10^{12}$ parameters/s and compute throughput $TP=2\times10^{13}$ FLOPS. Time is discretized with step size $\Delta=1$ s and tasks are sampled as $\overline{K}=1$.

Training-compute scaling enters through the parametric fit of Hoffmann et al.~\cite{Hoffmann2022}, yielding $\mathcal{L}_{\mathrm{irr}}=1.69$, $\gamma\approx0.34$, and $\Gamma\approx900$, implied by the parameters $E=1.69$, $A=406.4$, $B=410.7$, $\alpha=0.34$, and $\beta=0.28$, with $\gamma=\alpha$ and $\Gamma=A(1+\alpha/\beta)$.

Inference compute is modeled, as described in the main text, with skill-level success probabilities using a sigmoid model with parameter $\mathfrak{b}=5$. Each task consists of $m=50$ skills ($\omega=20$ tokens per skill attempt) and is parameterized by difficulty $l\in[1.7,1.9]$, linearly spaced across ten tasks. The error tolerance is fixed to $\varepsilon=0.1$ for all tasks. One task has a strict deadline permitting only a single feasible model, whereas two tasks have relaxed deadlines exceeding the maximum completion time of either model.

The average renewable energy budget $\overline{R}$ is chosen to be critical at $\overline{C}_{\mathrm{LB}}\approx593.5$. Renewable energy arrivals are sampled from a Gamma distribution with variance set to approximately that of the $\overline{C}_{\mathrm{LB}}\approx3.96\times 10^5$ ($\mathrm{Var}(R)\approx4\times10^{5}$).

Figure~\ref{fig:consumption_vs_latency} reports latency $\tau$ and total energy consumption for both model sizes as task difficulty increases from $(l,m)=(1.7,50)$ to $(1.9,50)$. This corresponds to inference token usage $\Omega$. Specifically, at $(l,m)=(1.7,50)$ we obtain $\tau=[34,24]$, energy $=[1023.3,946.9]$, and $\Omega=[57855,7841]$, while at $(l,m)=(1.9,50)$ we obtain $\tau=[9,11]$, energy $=[311.0,428.6]$, and $\Omega=[21967,3561]$, corresponding to the small and large models, respectively.

For Fig.~\ref{fig:auxiliarycostvstime}, regime transitions were detected via Bayesian Information Criterion (BIC) model selection. For each candidate breakpoint $T_k$, we fit a segmented model with square-root scaling ($m_1\sqrt{T} + b_1$) for $T < T_k$ and linear scaling ($m_2 T + b_2$) for $T \geq T_k$, comparing against a pure square-root baseline. The BIC score $n\log(\text{MSE}) + \lambda p \log(n)$ penalizes model complexity, where $n$ is the number of observations and $p$ is the parameter count ($p=2$ for the pure model, $p=5$ for the segmented model); we set $\lambda = 2$ to favor parsimony. The segmented model was accepted only when its optimal BIC score fell below that of the pure square-root fit. Results are shown over $T \in [0, 10000]$ s with $\Ex[D_T] \in [0, 120]$ kJ. The standard deviation of $D_T$ across trials is generally on the same order of magnitude as the expectation demonstrating the possibly significant variation in the accrued deficit.
\end{appendices}

\end{document}